\documentclass[sigconf]{acmart}
\usepackage{latexsym}

\usepackage{amssymb}
\usepackage{amsmath}
\usepackage{amsthm}
\usepackage{booktabs}
\usepackage{enumitem}
\usepackage{graphicx}
\usepackage{color}

\usepackage{subcaption}
\usepackage{pifont}
\usepackage{booktabs}
\usepackage{xcolor}
\usepackage{multirow}
\usepackage{array}
\usepackage{algorithm}
\usepackage{algpseudocode}

\AtBeginDocument{%
  \providecommand\BibTeX{{%
    \normalfont B\kern-0.5em{\scshape i\kern-0.25em b}\kern-0.8em\TeX}}}

\setcopyright{acmlicensed}
\copyrightyear{2026}
\acmYear{2026}
\acmDOI{XXXXXXX.XXXXXXX}
\acmConference[WWW '26]{We Need a More Robust Classifier: Dual Causal Learning Empowers Domain-Incremental Time Series Classification}{April 13-17, 2026}{Dubai, United Arab Emirates}
\acmISBN{978-1-4503-XXXX-X/24/08}

\begin{document}

\title{We Need a More Robust Classifier: Dual Causal Learning Empowers Domain-Incremental Time Series Classification}

\author{Zhipeng Liu}
\affiliation{%
  \institution{School of Software, Northeastern University}
  \city{Shenyang}
  \country{China}}
\email{2310543@stu.neu.edu.cn}

\author{Peibo Duan$^*$}
\affiliation{%
  \institution{School of Software, Northeastern University}
  \city{Shenyang}
  \country{China}}
\email{duanpeibo@swc.neu.edu.cn}\thanks{Peibo Duan and Binwu Wang are Corresponding authors.}

\author{Xuan Tang}
\affiliation{%
  \institution{School of Software, Northeastern University}
  \city{Shenyang}
  \country{China}}
\email{2471477@stu.neu.edu.cn}

\author{Haodong Jing}
\affiliation{%
  \institution{Institute of Artificial Intelligence and Robotics, Xi'an Jiaotong University}
  \city{Xi’an}
  \country{China}}
\email{jinghd@stu.xjtu.edu.cn}

\author{Mingyang Geng}
\affiliation{%
  \institution{College of Computer Science and Technology, National University of Defense Technology}
  \city{Changsha}
  \country{China}}
\email{gengmingyang13@nudt.edu.cn}

\author{Yongsheng Huang}
\affiliation{%
  \institution{School of Software, Northeastern University}
  \city{Shenyang}
  \country{China}}
\email{2371447@stu.neu.edu.cn}

\author{Jialu Xu}
\affiliation{%
  \institution{School of Software, Northeastern University}
  \city{Shenyang}
  \country{China}}
\email{xujialu@mails.neu.edu.cn}

\author{Bin Zhang}
\affiliation{%
  \institution{School of Software, Northeastern University}
  \city{Shenyang}
  \country{China}}
\email{zhangbin@mail.neu.edu.cn}

\author{Binwu Wang$^*$}
\affiliation{%
  \institution{School of Software, University of Science and Technology of China}
  \city{Hefei}
  \country{China}}
\email{wbw2024@ustc.edu.cn
}
\renewcommand{\shortauthors}{Liu et al.}

\begin{abstract}

The World Wide Web thrives on intelligent services that rely on accurate time series classification, which has recently witnessed significant progress driven by advances in deep learning. However, existing studies face challenges in domain incremental learning. In this paper, we propose a lightweight and robust dual-causal disentanglement framework (DualCD) to enhance the robustness of models under domain incremental scenarios, which can be seamlessly integrated into time series classification models. Specifically, DualCD first introduces a temporal feature disentanglement module to capture class-causal features and spurious features. The causal features can offer sufficient predictive power to support the classifier in domain incremental learning settings. To accurately capture these causal features, we further design a dual-causal intervention mechanism to eliminate the influence of both intra-class and inter-class confounding features. This mechanism constructs variant samples by combining the current class’s causal features with intra-class spurious features and with causal features from other classes. The causal intervention loss encourages the model to accurately predict the labels of these variant samples based solely on the causal features. Extensive experiments on multiple datasets and models demonstrate that DualCD effectively improves performance in domain incremental scenarios. We summarize our rich experiments into a comprehensive benchmark to facilitate research in domain incremental time series classification. The source code is available on GitHub\footnote{https://github.com/ZhipengLiu75/DualCD}. 

\end{abstract}



\maketitle

\section{Introduction}


Time series data are ubiquitous in web applications, producing large volumes of signals across domains such as human activity recognition and medical diagnostics \cite{huang2025exploiting,liu2025towards,ma2023rethinking,xie2025multivariate,zhang2023mlpst,ma2025less}. In recent years, researchers have developed various time series analysis models to advance this field. Notably, deep learning approaches have become predominant due to their powerful nonlinear modeling capabilities \cite{mohammadi2024deep,timecma2025liu}. A deep learning-based TSC model typically comprises an advanced feature encoder paired with a relatively simple classifier, often implemented as a fully connected layer. Generally, feature encoders that integrate cutting-edge techniques are highly effective at capturing complex temporal features within time series data. In contrast, the classifier acts as a bridge, mapping the generated temporal features to the label space.

While such an architecture may be suitable in fixed-domain scenarios, real-world applications often involve the continual emergence of new domains over time \cite{hadsell2020embracing,wang2024comprehensive}. Although these domains share a common label space, their data distributions may differ significantly. As a result, TSC models are expected to exhibit continual adaptability to incremental domains. However, they remain susceptible to the challenge of \textit{catastrophic forgetting} \cite{nguyen2019toward}, wherein performance on samples from previous domains degrades substantially as the model learns new domain knowledge.

Domain-incremental continual learning has emerged as a mainstream paradigm for addressing this challenge, and has been extensively studied in areas such as computer vision \cite{wang2024multi,luo2025domain}. The goal of this paradigm is to enable models to not only acquire new knowledge as they are continually exposed to novel domain data, but also retain previously learned information. Existing methods can generally be classified into two main categories. The first category comprises rehearsal-based methods, which “rehearse” old knowledge by replaying representative samples from previous domains \cite{rolnick2019experience}. This alleviates the forgetting of previously learned information during the acquisition of new knowledge. However, these methods incur significant storage overheads in practice and raise potential data privacy issues. Thus, increasing attention has shifted towards rehearsal-free methods. These approaches mitigate catastrophic forgetting without retaining data from previous tasks, offering stronger practicality and privacy preservation. For instance, some studies train domain-specific prompts and classifiers for each domain, then select the most appropriate prompt and classifier at inference phase based on the affinity between a given sample and each domain \cite{wang2022learning,wang2022s,smith2023coda}. While this strategy enhances adaptability, training multiple independent models introduces additional parameter overhead and restricts knowledge sharing and transfer across domains. To overcome these limitations, recent research has focused on learning class-general features that can serve as the basis for class discrimination and remain stable as new domains are introduced \cite{wang2024importance}. Prototype-based methods are a representative example \cite{wang2024non,wang2025dualcp}. However, such methods typically represent class-general features using a shared parameter space. Due to interference from the following two factors, models may end up capturing spurious general features:

\ding{182} \textit{Intra-class interference}. Although time series samples may share the same class labels, the underlying feature distributions can differ significantly across different domains. For example, in human activity recognition tasks, when learning “running” features from older adults (old domain), the model may regard lower step frequencies and smaller stride lengths as the general features of the “running” class. In contrast, when learning “running” features from younger adults (new domain), the model may perceive higher step frequencies and longer strides as the general features. \ding{183} \textit{Inter-class interference}. Time series samples from different classes may also exhibit highly similar feature representations. For instance, in human activity recognition, the features of low step frequency and low acceleration shown in “running” by older adults may be very similar to the features of “walking” by younger people. In such cases, a model trained well on samples from older adults may, after being fine-tuned with data from younger people (new domain), mistakenly associate low step frequency and low acceleration features with the “walking” class.

To extract discriminative class features in the context of domain-incremental scenarios, we propose a lightweight \textbf{Dual} \textbf{C}ausal \textbf{D}isen-tanglement framework (\textbf{DualCD}), which is designed to identify class-general causal features. These causal features capture the essential characteristics that distinguish one class from others and possess strong predictive power with respect to class labels. For example, in human activity recognition, features such as periodically accelerating movements serve as causal indicators for the “running” class, clearly differentiating it from “walking”, whereas features like step frequency and stride length may act as spurious features and should not be used as decisive class criteria.

Specifically, DualCD can be seamlessly integrated into existing time series classification models. The key philosophy underpinning DualCD is to guide the model in making predictions solely based on causal features rather than spurious ones, thereby ensuring robust performance in domain-incremental scenarios. To accurately identify causal features, DualCD incorporates a feature disentanglement module and a dual causal intervention mechanism. The disentanglement process employs two orthogonal masking matrices to partition the overall temporal representations into causal and spurious components. Drawing inspiration from causal theory \cite{neuberg2003causality,pearl2016causal,wang2023out}, we further introduce a dual intervention strategy. By sampling and recombining intra-class and inter-class variation patterns, this strategy generates multiple intervention samples, fostering the learning of high-quality causal features. For evaluation, we introduce a novel metric to comprehensively assess model performance in domain-incremental tasks. We validate our approach on multiple datasets with 12 advanced methods. Experimental results demonstrate that our method achieves SOTA performance. Moreover, by integrating DualCD with various mainstream time series classification models, we further verify its broad applicability. The contributions can be summarized as follows:

\begin{itemize}
    \item \textit{\textbf{Pioneer}}. To the best of our knowledge, this is the first study to address domain-incremental time series classification (DI-TSC), a setting that more accurately reflects real-world scenarios.
    
    \item \textit{\textbf{Method}}. We propose DualCD, which is specifically tailored for domain-incremental time series classification. DualCD combines orthogonal disentanglement with a dual causal intervention mechanism to extract causal features that remain robust in domain-incremental scenarios.
    
    \item \textit{\textbf{Benchmark}}. We establish a domain-incremental benchmark DI-TSC for time series to promote research in this field, which includes a novel evaluation metric, multiple datasets, and comprehensive baselines.

    \item \textit{\textbf{Experiment}}. We conduct extensive experiments on multiple benchmark datasets to evaluate the proposed DualCD. The experimental results demonstrate that DualCD consistently achieves superior performance, verifying its effectiveness and broad applicability.

\end{itemize}

\section{Related Work}

\subsection{Time Series Representation Learning}

Recent advances in deep learning have sparked significant interest in addressing core challenges in time series analysis, spanning tasks such as classification, forecasting, and anomaly detection \cite{shao2025hyperd,xia2025timeemb,sun2025ppgf,ma2025mobimixer,zhang2023autostl,sun2025hierarchical,li2026posterior,timecma2025liu,wang2025agentic}. Existing approaches primarily adopt two complementary perspectives to enhance predictive capabilities. On one hand, selecting appropriate backbone architectures to capture temporal representations is crucial. For instance, CNNs effectively model local dependencies due to their localized receptive fields \cite{wu2022timesnet,wang2023micn}; RNNs excel at capturing sequential dependencies through recurrent connections \cite{kong2025unlocking,liu2025dismsts};  while GNNs focus on modeling complex inter-variable relationships by capturing dependencies among multiple variables \cite{wang2024stone,liu2025attributed,wang2023pattern,liu2024dynamic,xue2025gated,liu2025distillation}. Recently, Transformers have emerged as a dominant architecture for these tasks \cite{ding2025timemosaic,liu2025timeformer,ma2025mofo,zhang2024skip,zhang2025probabilistic,wang2025correctformer}, owing to their powerful capability in modeling long-range dependencies. On the other hand, these models often incorporate specialized designs, such as multi-scale analysis \cite{wang2024timemixer}, periodicity, trend modeling \cite{wu2021autoformer}, and channel independence \cite{PatchTST}. However, these models typically assume single-task scenarios and suffer from catastrophic forgetting in continual learning environments involving multiple incremental tasks, leading to performance degradation.

\subsection{Continual Learning}

Continual learning aims to equip the model with the ability to continuously acquire knowledge across evolving tasks, without suffering from catastrophic forgetting. It is commonly studied under three scenarios: task-incremental (TIL), class-incremental (CIL), and domain-incremental learning (DIL) \cite{hsu2018re,van2019three}. TIL assumes task identities are available during both training and inference, which constrains its practicality in real-world scenarios. In CIL, classes generally originate from the same domain, thereby simplifying the continual learning process \cite{wei2023online,yan2024orchestrate}. We focus on DIL in this study, where class labels remain fixed, but domains differ markedly across tasks, and task identity is unavailable during inference. Existing methods for DIL can be broadly categorized into rehearsal-based \cite{zhang2022simple,jiang2025dupt} and rehearsal-free \cite{lamers2023clustering,wang2025dualcp} approaches.

\subsubsection{Rehearsal-based DIL}
Rehearsal strategies that utilize samples from a memory buffer or a generative model have proven to be effective in mitigating forgetting \cite{rebuffi2017icarl,chaudhry2019tiny,davalas2024rehearsal,yan2025rehearsal}. However, the long-term storage of training data in real-world machine learning applications raises critical concerns related to data privacy and substantial memory consumption. Moreover, training and deploying generative models require greater computational resources. In addition, generative models tend to memorize and reproduce sensitive information, raising potential legal and ethical risks of data misuse. This motivates us to focus on rehearsal-free methods.

\subsubsection{Rehearsal-free DIL}

Several representative rehearsal-free methods primarily rely on prompt learning, which typically involves learning a limited set of prompts and classifiers for each domain \cite{wang2022learning,wang2022s,smith2023coda,li2024personalized,wang2024non,luo2025domain}. However, these methods are often difficult to optimize, and their performance may vary non-monotonically with the number of trainable parameters. Therefore, some emerging methods aim to learn a unified classifier by capturing common information across multiple domains. For example, prototype-based approaches are used to enhance the compactness of feature representations \cite{wang2024non,wang2025dualcp}. However, they overlook both intra-domain and inter-domain interference in incremental continual learning, particularly for time series data. Consequently, adapting them to time series tasks is non-trivial.

\begin{table}[t]
\centering
\caption{Evaluation metrics for catastrophic forgetting in domain-incremental learning, including AF, RF, and our proposed PRF. \textit{The element $a_j^i$, at row $i$ and column $j$, represents the test accuracy on domain $j$ (column) after training the model on domains 1 through $i$ (row).} $\downarrow$ indicates that lower values correspond to less forgetting. Some models exhibit strong performance while maintaining low AF and RF scores.} 
\Huge
\renewcommand{\arraystretch}{1.4} 
\resizebox{0.48\textwidth}{!}{%
\begin{tabular}{c|ccc|ccc}

$a^i_j$ & $\mathcal{T}_1$ & $\mathcal{T}_2$ & $\mathcal{T}_3$ & AF($\downarrow$) & RF($\downarrow$) & PRF($\downarrow$) \\
\midrule
$\mathcal{T}_1$ 
& \textcolor[HTML]{FFBF08}{\textbf{0.8}}/\textcolor[HTML]{5599B5}{\textbf{0.4}} 
& - 
& - 
& - 
& - 
& - \\

$\mathcal{T}_2$ 
& \textcolor[HTML]{FFBF08}{\textbf{0.6}}/\textcolor[HTML]{5599B5}{\textbf{0.28}} 
& \textcolor[HTML]{FFBF08}{\textbf{0.9}}/\textcolor[HTML]{5599B5}{\textbf{0.3}} 
& - 
& \textcolor[HTML]{FFBF08}{\textbf{0.2}}/\textcolor[HTML]{5599B5}{\textbf{0.12}} 
& \textcolor[HTML]{FFBF08}{\textbf{0.25}}/\textcolor[HTML]{5599B5}{\textbf{0.3}} 
& \textcolor[HTML]{FFBF08}{\textbf{0.1}}/\textcolor[HTML]{5599B5}{\textbf{0.136}} \\

$\mathcal{T}_3$ 
& \textcolor[HTML]{FFBF08}{\textbf{0.4}}/\textcolor[HTML]{5599B5}{\textbf{0.22}} 
& \textcolor[HTML]{FFBF08}{\textbf{0.45}}/\textcolor[HTML]{5599B5}{\textbf{0.17}} 
& \textcolor[HTML]{FFBF08}{\textbf{0.6}}/\textcolor[HTML]{5599B5}{\textbf{0.4}} 
& \textcolor[HTML]{FFBF08}{\textbf{0.425}}/\textcolor[HTML]{5599B5}{\textbf{0.155}} 
& \textcolor[HTML]{FFBF08}{\textbf{0.5}}/\textcolor[HTML]{5599B5}{\textbf{0.421}} 
& \textcolor[HTML]{FFBF08}{\textbf{0.28}}/\textcolor[HTML]{5599B5}{\textbf{0.338}} \\

\end{tabular}
}

\label{metrics_Compare1}
\end{table}

\section{Preliminaries}
\subsection{Problem Statement}
In this study, we use $\mathbb{D} = \{\mathcal{D}^1, \mathcal{D}^2, \cdots, \mathcal{D}^T\}$ to denote a sequence of $T$ time series domains, where $\mathcal{D}^{t\in T}=\{\mathbf{X}_n^t,\mathbf{Y}^t_n\}_{n=1}^{N_t}$ represents the $t$-th domain including $N_t$ training samples. Samples from different domains follow different distributions, while sharing the same label space. $\mathbf{X}_n^t \in\mathbb{R}^{L\times M}$ indicates the $n$-th sample, where $L$ and $M$ denote the time series length and the number of feature channels. $\mathbf{Y}^t_n$ is the corresponding class label annotation.

Domain-Incremental Time Series Classification (\textbf{DI-TSC}) aims to continually train a model $f$ on a sequence of streaming domains $\mathbb{D}$, enabling accurate prediction on the new domain $\mathcal{D}^t$ while retaining performance on old ones $\{\mathcal{D}^i\}_{i=1}^{t-1}$.

\begin{figure*}[t]
  \centering
  \includegraphics[scale=0.8]{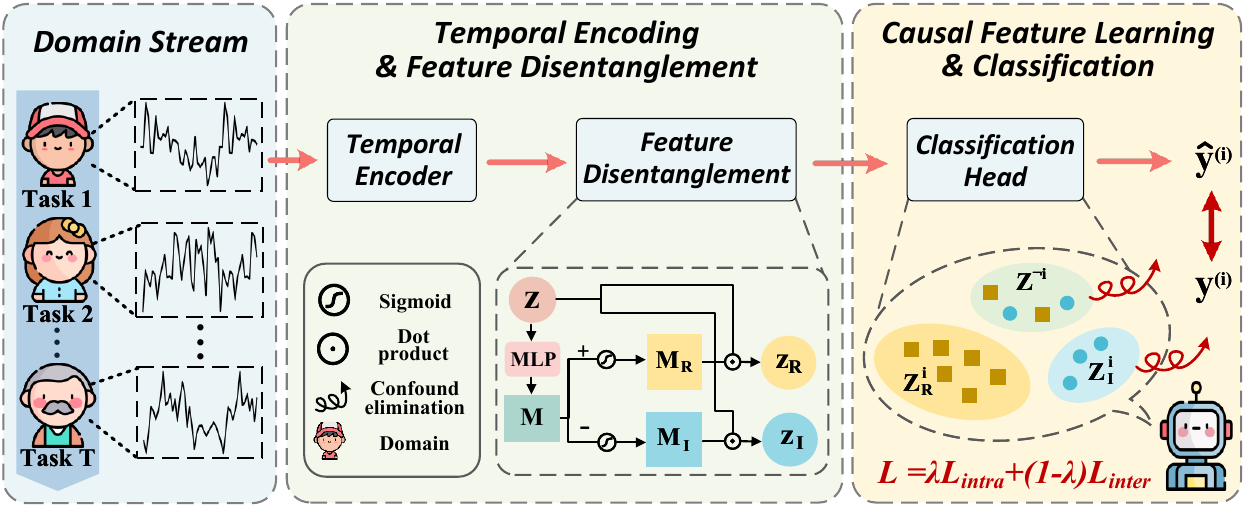}
  \caption{The framework of DualCD, which includes a temporal disentanglement module and a dual causal intervention mechanism, aims to learn class-invariant causal features.}
  \label{framework}
\end{figure*}

\subsection{Catastrophic Forgetting}
Catastrophic forgetting is a key challenge in domain-incremental learning, where a model forgets previously acquired knowledge while continuously learning from new domains. As illustrated in the left part of Table \ref{metrics_Compare1}, a toy example shows that Models A (yellow) and B (blue) suffer performance degradation on earlier domains as they learn from new ones. The element \( a_{j}^i \) at row \( i \) and column \( j \) denotes the accuracy on the test set of domain \( j \), after training the model from domain 1 to \( i \).

To evaluate a model’s effectiveness against catastrophic forgetting, two existing metrics are commonly used: Absolute Forgetting (AF) \cite{wang2024comprehensive} and Relative Forgetting (RF) \cite{wang2024improving}. Their formulas are as follows:

\begin{equation}
    \begin{aligned}
        \text{AF}&=\frac{1}{T-1}\sum_{j=1}^{T-1}\left(\displaystyle \max_{i \in \{1,2,\ldots,T-1\}} a^i_j-a^T_j\right), \\
        \text{RF}&=\frac{1}{T-1}\sum_{j=1}^{T-1}\left(\displaystyle \max_{i \in \{1,2,\ldots,T-1\}} \frac{(a^i_j-a^T_j)}{a^i_j} \right), 
    \end{aligned}
\end{equation}

\noindent AF and RF measure the absolute or relative difference between the highest previous accuracy on a domain and its current accuracy. However, as shown in the right part of Table \ref{metrics_Compare1}, the blue model achieves better (lower) AF and RF scores than the yellow model, which can be attributed to its initially low performance leaving limited room for further degradation on old domains.

To more accurately evaluate knowledge consolidation, overall model performance should be incorporated into the metric system. Accordingly, we propose a novel metric Performance-aware Relative Forgetting (PRF):
\begin{equation}
    \text{PRF}=\frac{1}{T-1}\sum_{j=1}^{T-1}\left(\displaystyle \max_{i \in \{1,2,\ldots,T-1\}} (1-a^T_j)\frac{(a^i_j-a^T_j)}{a^i_j} \right), 
\end{equation}
\noindent where PRF introduces a modulation factor based on the model’s overall performance, assigning a smaller penalization coefficient to models with higher accuracy. In this way, PRF not only reflects the effectiveness of a model in consolidating previously learned knowledge, but also takes the model’s overall performance into account.

\section{Method}

\subsection{Overall Framework}
Figure \ref{framework} and Algorithm \ref{alg} present the overall architecture of our DualCD, which can be seamlessly integrated into existing time series classification models to enable efficient domain continual learning. DualCD employs a dual causal disentanglement mechanism to extract class-invariant causal features that remain robust across dynamically changing domains. Specifically, the model first encodes input time series samples, then passes the resulting representations through a feature disentanglement module that separates causal and spurious components. Finally, two regularization terms, based on a dual causal intervention mechanism, are introduced to further enhance the representativeness of the causal features.

\subsection{Time Series Modeling and Classification} 
Recent research in time series classification has centered on designing increasingly sophisticated models to better capture the diverse sequential features of time series data. Many studies draw inspiration from computer vision and signal processing, employing techniques such as masked autoencoders (MAE) \cite{PatchTST}, multi-scale feature extraction \cite{Timemixer++,liu2025dismsts}, and Fourier transform-based approaches \cite{ma2025mobimixer}. These advancements have significantly improved the representational power of time series models. Importantly, our proposed DualCD framework is model-agnostic. Accordingly, we use the notation $\psi\left(\cdot\right)$ to refer to any time series classification model, such as DLinear \cite{zeng2023DLinear}, PatchTST \cite{PatchTST}, or other advanced architectures for modeling temporal features, followed by a classifier $\phi\left(\cdot\right)$ for final prediction. The overall process of label prediction can be formulated as follows:
    \begin{equation}   \mathbf{Z}^t=\psi(\mathbf{X}^t;\Theta_1^t),\widehat{\mathbf{Y}}^t=\phi\left(\mathbf{Z}^t;\Theta_2^t\right),
\end{equation}
where $\mathbf{X}^t\in\mathbb{R}^{L\times M}$ is the input time series, $\mathbf{Z}^t\in\mathbb{R}^{D_{\text{model}}}$ denotes the temporal representations generated by the time series classification model. $\widehat{\mathbf{Y}}^t$ is the predicted label. $\Theta_1^t$ and $\Theta_2^t$ are the parameters of $\psi(\cdot)$ and $\phi(\cdot)$ in the $t$-th domain, respectively. For the initial domain $t=1$, the parameters of all models are randomly initialized. In the subsequent continual domains $t >1$, we directly load the parameters from the previous domain for fine-tuning.

\subsection{Dual Causal Disentanglement for DI-TSC}

In domain-incremental scenarios, time series classification models often suffer from catastrophic forgetting when learning new domains. To mitigate this, we propose a dual causal disentanglement module that extracts class-specific causal features from temporal representations. These features remain robust across domains. The module consists of two key components: an orthogonal disentanglement module and a dual causal intervention mechanism.

\subsubsection{Orthogonal Disentanglement Strategy}
Existing models typically utilize a unified embedding $\mathbf{Z}$ \footnote{For simplicity, we omit the domain identifier $t$ in the subsequent sections.} to represent complex temporal features, assuming that these features can serve as an effective basis for category discrimination. However, this strategy encounters significant challenges in domain-incremental learning. On one hand, $\mathbf{Z}$ may alternately encode domain-specific features from newly encountered domains, thereby impairing the model’s ability to distinguish samples from previous domains. On the other hand, $\mathbf{Z}$ might be interpreted as representing generic features of incorrect classes within the new domain, leading to the misclassification of original samples from earlier domains. To address these challenges, we propose an orthogonal disentanglement module that distills class-specific causal features. These causal features are defined as those that are inherently distinct from features of other classes and exhibit strong predictive power for their corresponding labels.

Specifically, orthogonal disentanglement strategy employs a linear mapping to generate two orthogonal mask matrices $\mathbf{M}_{R}$ and $\mathbf{M}_{I}$, and then they are subsequently applied to the temporal representations for disentanglement. The orthogonality of these matrices ensures that the resulting feature sets are non-overlapping. The calculation process is formulated as follows:

\begin{equation}
    \label{decompose}
    \begin{aligned}
        \mathbf{M}&=\text{MLP}(\mathbf{Z}),\\
        \mathbf{M}_{R}&={\text{Sigmoid}}(\mathbf{M}),\\
        \mathbf{M}_{I}&={\text{Sigmoid}}(-\mathbf{M}),\\
        \mathbf{Z}_{R}&=\mathbf{M}_{R}\odot\mathbf{Z}, \mathbf{Z}_{I}= \mathbf{M}_{I}\odot\mathbf{Z},
        \end{aligned}
\end{equation} 

\noindent where a single-layer MLP is used to generate an intermediate score vector $\mathbf{M}\in\mathbb{R}^{D}$ from $\mathbf{Z}$. We then utilize sigmoid activations to $\mathbf{M}$ and $-\mathbf{M}$ to obtain a pair of negatively correlated soft masks: $\mathbf{M}_R$ and $\mathbf{M}_I$. Finally, the causal feature \(\mathbf{Z}_R\in\mathbb{R}^{D}\) and the spurious feature \(\mathbf{Z}_I\in\mathbb{R}^{D}\) are explicitly disentangled by element-wise multiplication of \(\mathbf{Z}\) with \(\mathbf{M}_R\) and \(\mathbf{M}_I\), respectively.

\subsection{Dual Causal Intervention Perturbation}

\subsubsection{Causal Theory} Causal theory posits that effective causal features should retain sufficient predictive power even in changing environments \cite{zhang2022dynamic}. Its objective can be formulated as,
\begin{equation}
\begin{array}{r}
\min_{\Theta_1 , \Theta_2 } \mathbb{E}_{\left(\mathbf{X} , \mathbf{Y} \right) \sim p\left(\mathbb{X} , \mathbb{Y} \right)} \mathcal{L}\left(\phi\left(\psi(\mathbf{X} ;\Theta_1 ); \Theta_2 \right), \mathbf{Y} \right)+ \\
\lambda \operatorname{Var}\left(\mathbb{E}_{\left(\mathbf{X} , \mathbf{Y} \right) \sim p\left(\mathbb{X} , \mathbb{Y}  \mid \operatorname{do}\left(\mathbf{Z}_{I}=s\right)\right)} \mathcal{L}\left(\phi\left(\psi(\mathbf{X} ;\Theta_1 ); \Theta_2 \right), \mathbf{Y} \right)\right),
\end{array}
\end{equation}

\noindent where `do' denotes do-calculas to intervene the original distribution \cite{pearl2016causal}, $s \in \mathbb{S}$ is the intervention and $\lambda$ is a balancing hyperparameter. In plain terms, given the causal feature, the confounding factor has no effect on the label $\mathbf{Y}$. Thus, intervening on $\mathbf{Z}_{I}$ while keeping $\mathbf{Z}_{R}$ fixed should not alter the prediction of $\mathbf{Y}$.

\begin{algorithm}[t]
\caption{DualCD for Domain-Incremental Time Series Classification} 
\label{alg}
\begin{algorithmic}[1]  
\State \textbf{Require:} Domain-incremental time series dataset $\mathbb{D} = \{\mathcal{D}^1, \mathcal{D}^2, \cdots, \mathcal{D}^T\}$, training epochs $E$.
\For{$t=\{1,2,\cdots,T\}$}
    \If{$t = 1$}
        \State Randomly initialize the model parameters.
    \Else
        \State Load the model trained in the ($t-1$)$_{th}$ domain.
    \EndIf
    \For{$e=\{1,2,\cdots,E\}$}
    \State Generate temporal representation $\mathbf{Z}^t$.
    \State Feature disentanglement to obtain $\mathbf{Z}_R^t$ and $\mathbf{Z}_I^t$.
    \State Introduce intra-class causal perturbation as Eq.\eqref{intra_perturbation}.
    \State Introduce inter-class causal perturbation as Eq.\eqref{inter_perturbation}.
    \State Model optimization as Eq.\eqref{complete_loss}. 
    \EndFor
    \State Save the optimized parameters of the model.
\EndFor
\end{algorithmic}
\end{algorithm}

Essentially, perturbing $\mathbb{S}$ implicitly creates multiple variant samples, and these perturbations allow the model to capture robust causal features. However, directly intervening in these patterns faces significant complexity challenges \cite{zhang2022dynamic}. Moreover, the quality of the learned causal features highly depends on the design of the intervention strategy. Thus, we introduce a dual causal intervention mechanism designed for domain-incremental time series classification, which helps to eliminate confounding factors within and across classes, thereby generating high-quality causal features, with further details provided in the following sections.

\subsubsection{Intra-Class Causal Perturbation}

We randomly select spurious features from other intra-class samples to replace the original spurious features. Specifically, the perturbation process is defined as,
\begin{equation}
\label{intra_perturbation}
\mathbf{Z}_{R}^{(i)}, \mathbf{Z}_{I}^{(i)}  \leftarrow  \mathbf{Z}_{R}^{(i)}, \mathbf{Z}_{I}^{(i^{'})},
\end{equation}

\noindent where $\mathbf{Z}_{I}^{(i^{'})}$ denotes a randomly sampled spurious feature from other samples belonging to the $i$-th class. Subsequently, we combine $\mathbf{Z}_{R}^{(i)}$ and $\mathbf{Z}_{I}^{(i)}$ via element-wise addition to generate the output $\widetilde{\mathbf{Z}}^{(i)}_{\text{intra}}$, which is then fed into the classifier to generate the predicted label. Since our assumption is that the model relies solely on causal features, the label for this variant sample remains unchanged. We then compute the following intra-class causal perturbation loss:
\begin{equation}
    \mathcal{L}_{\text{intra}} =  
        \mathrm{CE}\left( \phi(\widetilde{\mathbf{Z}}^{(i)}_{\text{intra}};\Theta_2), \mathbf{y}^{(i)} \right),
\end{equation}

\noindent where $\mathrm{CE}(\cdot, \cdot)$ denotes the cross-entropy loss.

\subsubsection{Inter-Class Causal Perturbation}
To generate causal features that are clearly distinguishable from those of other classes, we randomly sample the causal features of a sample from any other class and use them to replace the original spurious features. This perturbation process can be denoted as:
\begin{equation}
\label{inter_perturbation}
\mathbf{Z}_{R}^{(i)}, \mathbf{Z}_{I}^{(i)}  \leftarrow  \mathbf{Z}_{R}^{(i)}, \mathbf{Z}_{R}^{(\neg i)},
\end{equation}

\noindent where $\mathbf{Z}_{R}^{(\neg i)}$ indicates the causal features randomly sampled from a sample belonging to other classes. Then, we add $\mathbf{Z}_{R}^{(i)}$ and $\mathbf{Z}_{I}^{(i)}$, and define the result as $\widetilde{\mathbf{Z}}^{(i)}_{\text{inter}}$. Similarly, the label of this variant feature should remain unchanged. We then compute the inter-class causal perturbation loss, denoted as:

\begin{equation}
    \mathcal{L}_{\text{inter}} =  
        \mathrm{CE}\left( \phi(\widetilde{\mathbf{Z}}^{(i)}_{\text{inter}};\Theta_2), \mathbf{y}^{(i)} \right).
\end{equation}

\subsubsection{Optimization Objective} Finally, the model is optimized based on the following loss function:
\begin{equation}
    \label{complete_loss}\mathcal{L}=\lambda\mathcal{L}_{\text{intra}}+(1-\lambda)\mathcal{L}_{\text{inter}},
\end{equation}
\noindent where the hyperparameter $\lambda\in [0,1]$ controls the trade-off between $\mathcal{L}_{\text{intra}}$ and $\mathcal{L}_{\text{inter}}$.

\begin{table}[tbp]
      \caption{Dataset statistics.}
  \renewcommand{\arraystretch}{.5}
  \centering
    \resizebox{1.\linewidth}{!}{
    \begin{tabular}{ccccccccc}
    \toprule[1.pt]
\textbf{Dataset}&\textbf{HAR}&\textbf{HHAR}&\textbf{ISRUC-S3}&\textbf{Sleep-EDF}\\
    \toprule[.7pt]
    \#Train &7,194&10,336&6,007&29,606\\
    \#Val   &1,550&2,214&1,299&6,349\\
    \#Test  &1,555&2,222&1,283&6,353\\
    Length&128&128&3,000&3,000\\
    \#Variable&9&3&10&1\\
    \#Class&6&6&5&5\\
    \#Domain&10&9&10&10\\
    \bottomrule[1.pt]
    \end{tabular}}%
  \label{dataset}
\end{table}%

\begin{table*}[t]
\large
\caption{Performance comparison with baselines. \underline{Underline} indicates the best performance within each family, while {\textbf{bold}} highlights the overall best results. * denotes that the improvement (e.g., ACC $\uparrow$) or decline (e.g., RF $\downarrow$ and PRF $\downarrow$) of our method, compared to all baselines, is significant, based on a t-test with a $p<0.01$.}
\renewcommand{\arraystretch}{.5}
  \centering
    \resizebox{1.0\linewidth}{!}{
    \begin{tabular}{cc|ccccccccccccc}
    \toprule[1.pt]

    &\multirow{2}{*}{Methods}&
    \multicolumn{3}{c}{HAR}&\multicolumn{3}{c}{HHAR}&\multicolumn{3}{c}{ISRUC}&\multicolumn{3}{c}{Sleep-EDF}\\
    
    \cmidrule(lr){3-5}\cmidrule(lr){6-8}\cmidrule(lr){9-11}\cmidrule(lr){12-14}

    &&ACC&RF&PRF&ACC&RF&PRF&ACC&RF&PRF&ACC&RF&PRF\\
    \toprule[.7pt]

    \multirow{9}{*}{TSC}&DLinear&0.4737&0.2894&0.1610&0.3268&0.5630&0.4231&0.2185&0.1720&0.1380&0.2555&\underline{0.1284}&0.0846\\

    \cmidrule(lr){2-14}
    &PatchTST&0.7946&0.1604&0.0386&0.5240&0.4932&0.2706&0.5755&0.2223&0.1053&0.7047&0.1380&0.0487\\

    \cmidrule(lr){2-14}
    &iTransformer&0.7959&0.1872&0.0390&0.6944&0.2673&0.1327&0.3280&0.1813&0.1295&0.4091&0.1368&0.0473\\

    \cmidrule(lr){2-14}
    &xPatch&0.7408&0.2246&0.0662&0.6739&0.2993&0.1477&0.4673&0.1642&0.1403&0.5937&0.1426&0.0625\\

    \cmidrule(lr){2-14}
    &TimeMixer++&0.7735&0.2066&0.0543&0.6484&0.3769&0.2009&0.5809&0.1715&0.0805&0.6681&0.1436&0.0506\\

    \cmidrule(lr){2-14}
    &PatchMLP&0.7521&0.1947&0.0564&0.6184&0.3828&0.1993&0.3712&0.1922&0.1333&0.4730&0.1339&0.0456\\
    
    \cmidrule(lr){2-14}
    &DisMS-TS&\underline{0.8240}&\underline{0.1577}&\underline{0.0337}&\underline{0.7298}&\underline{0.2576}&\underline{0.1276}&\underline{0.6676}&\underline{0.0988}&\underline{0.0583}&\underline{0.7171}&0.1306&\underline{0.0436}\\
    
    \cmidrule(lr){1-14}

    \multirow{6}{*}{DIL}
    &EWC&0.7903&0.1799&0.0436&0.6816&0.3287&0.1686&0.5755&0.2223&0.1053&0.6963&0.1559&0.0562\\
    \cmidrule(lr){2-14}
    &LwF&0.8171&0.1648&0.0378&0.7099&0.3034&0.1514&0.6171&0.1292&0.0596&0.7031&0.1454&0.0557\\
    \cmidrule(lr){2-14}
    &DT2W&0.8158&0.1667&0.0413&0.6872&0.3325&0.1585&0.6432&0.1295&0.0583&0.7010&0.1431&0.0490\\
    \cmidrule(lr){2-14}
    &DualPrompt&0.8262&0.1635&0.0372&0.7090&0.3054&0.1568&0.6397&0.1127&0.0662&0.7134&0.1425&0.0469\\
    \cmidrule(lr){2-14}
    &DualCP&\underline{0.8302}&\underline{0.1562}&\underline{0.0310}&\underline{0.7328}&\underline{0.2518}&\underline{0.1137}&\underline{0.6898}&\underline{0.0951}&\underline{0.0531}&\underline{0.7211}&\underline{0.1320}&\underline{0.0451}\\
    
    \cmidrule(lr){1-14}
    \textbf{Ours}&\textbf{DualCD} &\textbf{0.8565}$^*$&\textbf{0.1410}$^*$&\textbf{0.0266}$^*$&\textbf{0.7789}$^*$&\textbf{0.2044}$^*$&\textbf{0.0810}$^*$&\textbf{0.7277}$^*$&\textbf{0.0864}$^*$&\textbf{0.0318}$^*$&\textbf{0.7459}$^*$&\textbf{0.1264}$^*$&\textbf{0.0331}$^*$\\
    \bottomrule[1.pt]
    \end{tabular}}

    \label{performance}
\end{table*}

\section{Experiment}



\subsection{Experimental Settings}
\subsubsection{Datasets.}




We use four benchmark time series classification datasets covering human activity recognition and EEG prediction tasks: HAR, HHAR, ISRUC-S3, and sleep-EDF \cite{He_2023_DomainAdaptation,wang2024graph}. The HAR and HHAR datasets consist of sensor recordings from 30 and 9 users, respectively, while ISRUC-S3 and Sleep-EDF include data from 10 and 20 subjects. For HHAR and ISRUC-S3, we adopt the original subject-wise domain partitioning, treating each subject as an individual domain. For HAR and sleep-EDF, we construct domain-incremental settings by grouping data from selected subjects into 10 domains, ensuring that each domain contains data from a disjoint set of users. Detailed dataset statistics are shown in Table~\ref{dataset}.

\subsubsection{Evaluation Metrics.}
We use average accuracy (ACC $\uparrow$), average relative forgetting (RF $\downarrow$) \cite{wang2024improving}, and our proposed average performance-aware relative forgetting (PRF $\downarrow$) to measure the performance.

\subsubsection{Baselines.}

We compare DualCD against 12 baseline models, which fall into two families: \textbf{time series classification} and \textbf{domain-incremental continual learning}. The former includes seven time series classification models, DLinear \cite{zeng2023DLinear}, PatchTST \cite{PatchTST}, iTransformer \cite{iTransformer}, xPatch \cite{xPatch}, TimeMixer++ \cite{Timemixer++}, PatchMLP \cite{PatchMLP}, and DisMS-TS \cite{liu2025dismsts}. The second consists of five domain-incremental continual learning models, EWC \cite{EWC}, LwF \cite{li2017learning}, DT2W \cite{qiao2023class}, DualPrompt \cite{wang2022dualprompt}, and DualCP \cite{wang2025dualcp}. To ensure a fair comparison, we adopt the encoder of DisMS-TS as the backbone for both the domain-incremental methods and our DualCD.

\subsubsection{Implementation Details.}
All experiments are conducted on a single NVIDIA GeForce RTX 4090 GPU with 24 GB of memory. All models employ the Adam optimizer \cite{adam2014method} for training. For all datasets, each domain is split into training, validation, and test sets with proportions of 70\%, 15\%, and 15\%, respectively. To guarantee adequate learning for each domain, the best-performing model on the validation set is utilized to initialize the training of the subsequent domain. All experiments are independently conducted five times, and the average results are reported.

\begin{figure}[t]
  \centering
  \includegraphics[scale=0.26]{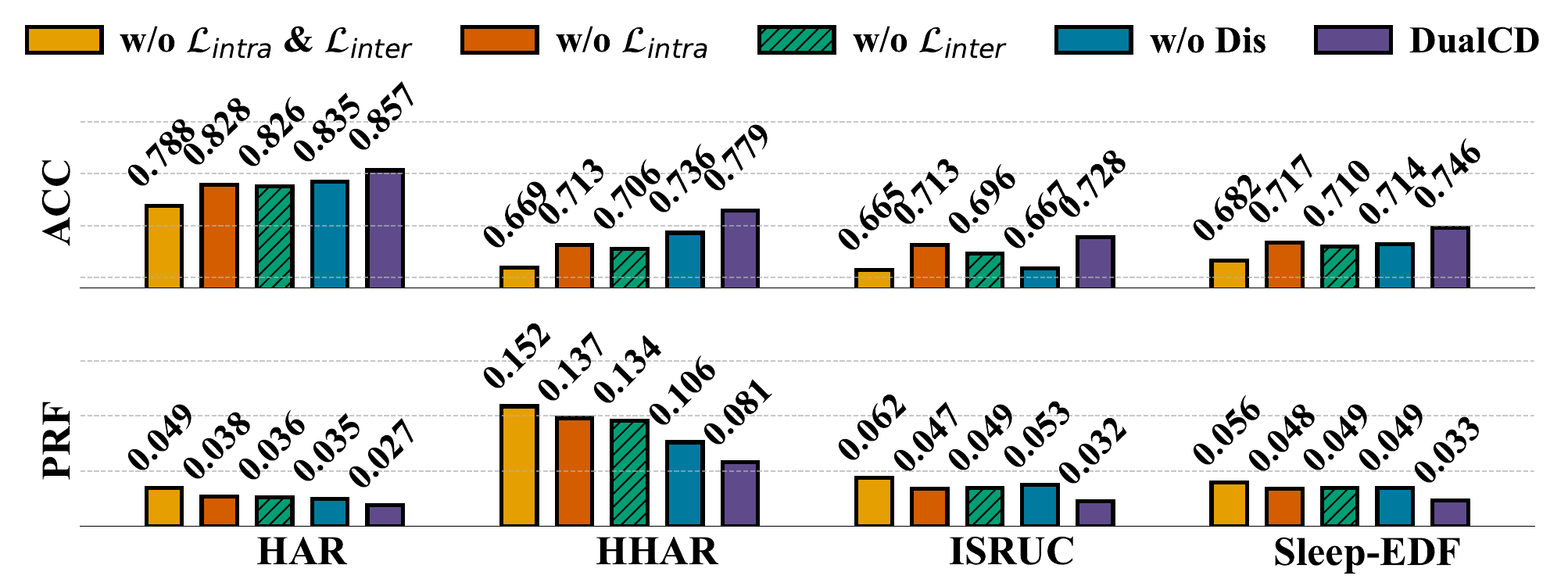}
  \caption{Ablation study on the four datasets.}
  \label{Ablation}
\end{figure}

\subsection{Overall Performance Comparison}

Table~\ref{performance} presents the comparison results, from which we can observe that our proposed DualCD achieves the best overall performance across the four datasets. Moreover, to validate the reliability of the proposed DualCD, we independently conduct five runs and perform a t-test with a significance level of $\alpha = 0.01$ to assess performance differences between DualCD and other baselines. The results show that DualCD significantly outperforms ($p < 0.01$) all baselines on four datasets. Specifically, on the one hand, DualCD outperforms the state-of-the-art time series classification baseline, DisMS-TS, with accuracy (ACC) improvements of 3.94\%, 6.72\%, 9.00\%, and 4.01\% across four datasets, respectively. Compared to the best-performing domain-incremental learning baseline, DualCP, our DualCD achieves gains of 3.16\%, 6.29\%, 5.49\%, and 3.43\%, respectively. On the other hand, our DualCD shows respective reductions of 10.58\%, 20.65\%, 12.55\%, and 3.52\% compared to DisMS-TS, and 9.73\%, 18.82\%, 9.14\%, and 4.24\% compared to DualCP, in terms of relative forgetting (RF). This can be attributed to our dual causal disentanglement classifier, which performs feature disentanglement followed by dual causal intervention mechanisms to jointly promote stability and plasticity. 

Moreover, we observe that DLinear achieves a superior RF compared to all other baselines on the Sleep-EDF dataset, which suggests minimal forgetting and seemingly better results. However, its low accuracy substantially limits its practical utility. In contrast, our proposed PRF provides a more comprehensive and reliable model evaluation, which offers greater guidance for model selection.

\begin{figure}[t]
    \centering
    \begin{subfigure}[b]{0.21\textwidth}
        \centering
    \includegraphics[width=\linewidth]{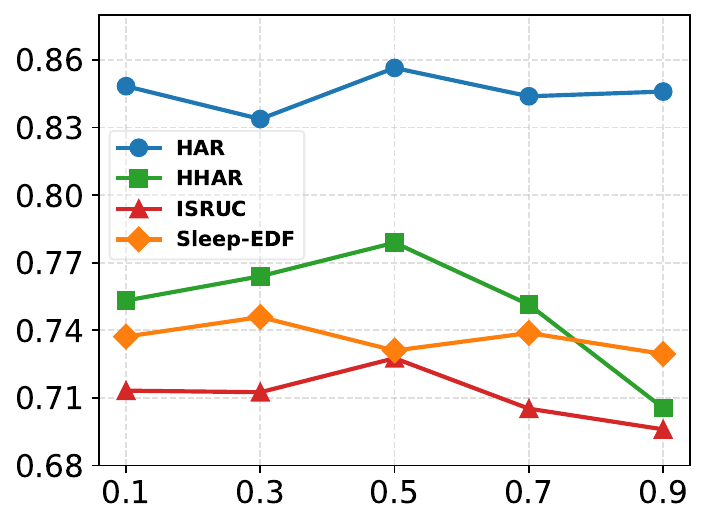}
        \caption{ACC.}
        \label{hyperacc}
    \end{subfigure} 
    \hspace{0.00\textwidth}
    \begin{subfigure}[b]{0.21\textwidth}
        \centering
        \includegraphics[width=\linewidth]{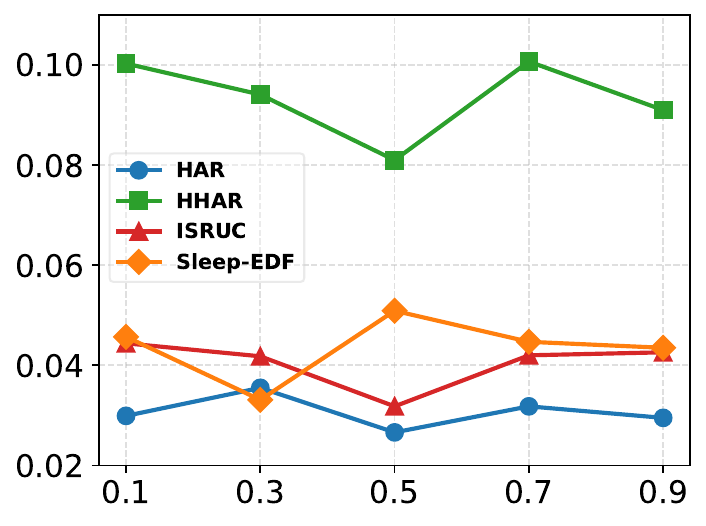}
        \caption{PRF.}
         \label{hyperprf}
    \end{subfigure}
    \caption{Hyperparameter analysis on the HAR dataset.} 
    \label{Hyperparameter}
\end{figure}

\subsection{Ablation Study}
To evaluate the contribution of each component, we conduct an ablation study by constructing the following four variants and comparing them with the complete DualCD.

\begin{itemize}
    \item `w/o $\mathcal{L}_{\text{intra}}$': The intra-class intervention is not considered. 
    \item `w/o $\mathcal{L}_{\text{inter}}$': The inter-class intervention is not considered.
    \item `w/o $\mathcal{L}_{\text{intra}}$\&$\mathcal{L}_{\text{inter}}$': Both intra- and inter-class interventions are not considered.
    \item `w/o Dis': The disentanglement module is replaced with two MLP modules. 
\end{itemize}

\noindent Ablation results on the four datasets, are presented in Figure~\ref{Ablation}. We can observe that each module contributes to higher ACC and lower PRF. Specifically, in terms of ACC, the four variants exhibit maximum degradations of 9.37\%, 8.47\%, 14.12\% and 8.37\%, respectively, when compared to the complete DualCD. Regarding the PRF, the four variants demonstrate increases of up to 93.75\%, 69.13\%, 65.43\%, and 65.62\%, respectively, when compared with the complete DualCD. These results collectively validate the effectiveness of our disentanglement module and dual causal intervention mechanism, underscoring the importance of capturing class-invariant causal features for domain-incremental time series classification.

\begin{table}[t]
\setlength{\tabcolsep}{3pt}
\caption{Compatibility analysis of DualCD integrated with more time series models.}
\renewcommand{\arraystretch}{.5}
  \centering
    \resizebox{1.0\linewidth}{!}{
    \begin{tabular}{c|c|cccc}
    \toprule[.7pt]
    
    \multirow{2}{*}{Dataset}&\multirow{2}{*}{Model}& \multicolumn{2}{c}{w/o DualCD}&\multicolumn{2}{c}{with DualCD}\\
    
    \cmidrule(lr){3-4}\cmidrule(lr){5-6}
    &&ACC($\uparrow$)&PRF($\downarrow$)&ACC($\uparrow$)&PRF($\downarrow$)\\
    \cmidrule(lr){1-6}

    \multirow{13}{*}{HAR}
    &DLinear&0.4737&0.1610&\textbf{0.7876}$_{\uparrow 66.26\%}$&\textbf{0.0353}$_{\downarrow 78.07\%}$\\
    \cmidrule(lr){2-6}
    
    &PatchTST&0.7946&0.0386&\textbf{0.8378}$_{\uparrow 5.43\%}$&\textbf{0.0297}$_{\downarrow 23.05\%}$\\
    \cmidrule(lr){2-6}
    
    &iTransformer&0.7959&0.0390&\textbf{0.8412}$_{\uparrow 5.70\%}$&\textbf{0.0249}$_{\downarrow 36.15\%}$\\ 
    \cmidrule(lr){2-6}
    
    &xPatch&0.7408&0.0662&\textbf{0.8009}$_{\uparrow 8.11\%}$&\textbf{0.0467}$_{\downarrow 29.45\%}$\\
    \cmidrule(lr){2-6}
    &TimeMixer++&0.7735&0.0543&\textbf{0.8295}$_{\uparrow 7.23\%}$&\textbf{0.0348}$_{\downarrow 35.91\%}$\\ 
    \cmidrule(lr){2-6}
     
     &PatchMLP&0.7521&0.0564&\textbf{0.8057}$_{\uparrow 7.12\%}$&\textbf{0.0302}$_{\downarrow 46.45\%}$\\
     
    \cmidrule(lr){1-6}
    
     \multirow{13}{*}{HHAR}
     &DLinear&0.3268&0.4231&\textbf{0.5937}$_{\uparrow 81.67\%}$&\textbf{0.2221}$_{\downarrow 47.51\%}$\\
    \cmidrule(lr){2-6}
    
    &PatchTST&0.5240&0.2706&\textbf{0.5753}$_{\uparrow 9.79\%}$&\textbf{0.2196}$_{\downarrow 18.85\%}$\\
    \cmidrule(lr){2-6}
    
    &iTransformer&0.6944&0.1327&\textbf{0.7356}$_{\uparrow 5.93\%}$&\textbf{0.1285}$_{\downarrow 3.17\%}$\\ 
    \cmidrule(lr){2-6}
    
    &xPatch&0.6739&0.1477&\textbf{0.7380}$_{\uparrow 9.51\%}$&\textbf{0.0889}$_{\downarrow 39.81\%}$\\
    \cmidrule(lr){2-6}
    
    &TimeMixer++&0.6484&0.2009&\textbf{0.7273}$_{\uparrow 12.17\%}$&\textbf{0.1509}$_{\downarrow 24.89\%}$\\ 
    \cmidrule(lr){2-6}
     
     &PatchMLP&0.6184&0.1993&\textbf{0.7380}$_{\uparrow 19.34\%}$&\textbf{0.0889}$_{\downarrow 55.39\%}$\\
    \cmidrule(lr){1-6}

     \multirow{13}{*}{ISRUC}
     &DLinear&0.2185&0.1380&\textbf{0.2502}$_{\uparrow 14.51\%}$&\textbf{0.1297}$_{\downarrow 6.01\%}$\\
    \cmidrule(lr){2-6}
    
    &PatchTST&0.5755&0.1053&\textbf{0.6138}$_{\uparrow 6.66\%}$&\textbf{0.0459}$_{\downarrow 56.41\%}$\\
    \cmidrule(lr){2-6}
    
    &iTransformer&0.3280&0.1295&\textbf{0.3647}$_{\uparrow 11.19\%}$&\textbf{0.0969}$_{\downarrow 25.17\%}$\\ 
    \cmidrule(lr){2-6}
    
    &xPatch&0.4673&0.1403&\textbf{0.5818}$_{\uparrow 24.50\%}$&\textbf{0.0582}$_{\downarrow 58.52\%}$\\
    \cmidrule(lr){2-6}
    
    &TimeMixer++&0.5809&0.0805&\textbf{0.6233}$_{\uparrow 7.30\%}$&\textbf{0.0468}$_{\downarrow 41.86\%}$\\ 
    \cmidrule(lr){2-6}
     
     &PatchMLP&0.3712&0.1333&\textbf{0.4976}$_{\uparrow 34.05\%}$&\textbf{0.0487}$_{\downarrow 63.47\%}$\\
    \cmidrule(lr){1-6}

     \multirow{13}{*}{Sleep-EDF}
     &DLinear&0.2555&0.0846&\textbf{0.4363}$_{\uparrow 70.76\%}$&\textbf{0.0440}$_{\downarrow 47.99\%}$\\
    \cmidrule(lr){2-6}
    
    &PatchTST&0.7047&0.0487&\textbf{0.7321}$_{\uparrow 3.89\%}$&\textbf{0.0350}$_{\downarrow 28.13\%}$\\
    \cmidrule(lr){2-6}
    
    &iTransformer&0.4091&0.0473&\textbf{0.4673}$_{\uparrow 14.23\%}$&\textbf{0.0434}$_{\downarrow 8.25\%}$\\ 
    \cmidrule(lr){2-6}
    
    &xPatch&0.5937&0.0625&\textbf{0.6450}$_{\uparrow 8.64\%}$&\textbf{0.0424}$_{\downarrow 32.16\%}$\\
    \cmidrule(lr){2-6}
    
    &TimeMixer++&0.6681&0.0506&\textbf{0.7159}$_{\uparrow 7.15\%}$&\textbf{0.0399}$_{\downarrow 21.15\%}$\\ 
    \cmidrule(lr){2-6}
     
     &PatchMLP&0.4730&0.0456&\textbf{0.5446}$_{\uparrow 15.14\%}$&\textbf{0.0426}$_{\downarrow 6.58\%}$\\
    \bottomrule[.7pt]
    \end{tabular}}
    \label{further_integrate}
\end{table}

\subsection{Hyperparameter Analysis}
To evaluate the effect of $\lambda$, which controls the trade-off between $\mathcal{L}_{\text{intra}}$ and $\mathcal{L}_{\text{inter}}$, we select its value from \{0.1, 0.3, 0.5, 0.7, 0.9\}. The experiments are conducted on four datasets. In terms of ACC, DualCD achieves its best performance at $\lambda = 0.5$ on HAR, HHAR, and ISRUC datasets, while the optimal value on the Sleep-EDF dataset is $\lambda = 0.3$. A similar observation holds in terms of the PRF. Extremely large or small values of $\lambda$ do not lead to optimal performance, which indicates that interventions at both intra-class and inter-class levels are essential for learning causal features. It is noted that $\lambda=0$ and $\lambda=1$ indicate the absence of $\mathcal{L}_{\text{intra}}$ and $\mathcal{L}_{\text{inter}}$, respectively, as analyzed in the ablation study.

\subsection{Plug-and-Play Compatibility Analysis}
We further integrate DualCD into other time series models, with the results on the four datasets shown in Table~\ref{further_integrate}. DualCD consistently improves both stability and plasticity across all models. Taking DLinear on the four datasets as an example, DualCD achieves up to 66.26\% improvement in ACC and up to 78.07\% reduction in PRF. This can be attributed to its linear structure, which limits its ability to capture rich temporal dependencies, making it more susceptible to learning superficial features that are spuriously correlated with class labels. Overall, the results further validate the effectiveness of DualCD in capturing causal features and highlight its potential as a plugin to enhance time series models in domain-incremental time series classification.

\begin{figure}[t]
    \centering
    \begin{subfigure}[b]{0.21\textwidth}
        \centering
    \includegraphics[width=\linewidth]{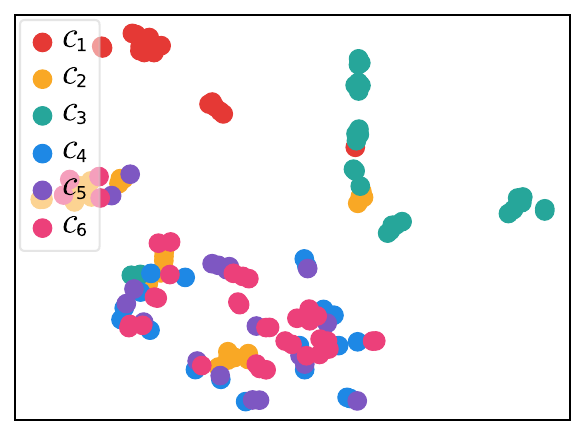}
        \caption{$\mathbf{Z}$ generated from the vanilla DLinear.}
        \label{Visualization_without}
    \end{subfigure} 
    \hspace{0.00\textwidth}
    \begin{subfigure}[b]{0.21\textwidth}
        \centering
        \includegraphics[width=\linewidth]{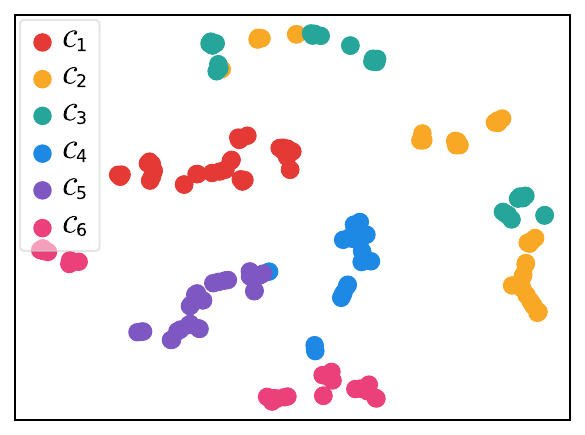}
        \caption{$\mathbf{Z}_R$ generated from DualCD-enhanced DLinear.}
         \label{Visualization_with}
    \end{subfigure}
    \caption{Visualization analysis conducted on the test set of domain 1 from the HAR dataset.} 
    \label{Visualization}
\end{figure}

\subsection{Analysis of Causal Representation $\mathbf{Z}_R$}

To intuitively assess the quality of the causal representations $\mathbf{Z}_R$, we compare them with the conventional temporal representations $\mathbf{Z}$, which are obtained from the original time series models without incorporating our DualCD.

\subsubsection{Visualization Analysis}

To assess whether these representations are discriminative, we take the DLinear model as a representative example. After training on all domains, we visualize $\mathbf{Z}$ and $\mathbf{Z}_R$ from the first domain using t-SNE. Figure~\ref{Visualization}-(a) presents the visualization of the temporal representation $\mathbf{Z}$ generated by the vanilla DLinear (without DualCD) in Equation (3). Figure~\ref{Visualization}-(b) shows the causal representation $\mathbf{Z}_R$ obtained by integrating DualCD as defined in Equation (4). We observe that the representations extracted by vanilla DLinear exhibit substantial overlap between classes, indicating limited discriminative ability. In contrast, the representations learned with DualCD display well-separated clusters and clearer decision boundaries. This demonstrates that DualCD effectively extracts class-invariant causal features, thereby providing a more robust and informative feature space for downstream classification.

\subsubsection{Cross-Domain Feature Difference Analysis}

\begin{figure}[t]
    \centering
    \begin{subfigure}[b]{0.23\textwidth}
        \centering
    \includegraphics[width=\linewidth]{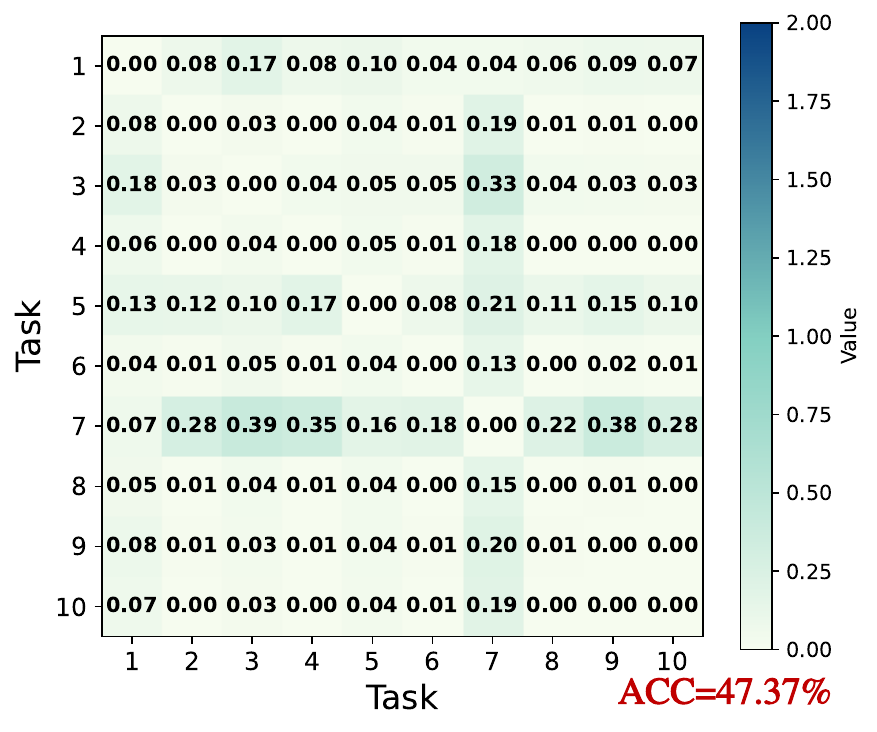}
        \caption{Difference matrix of $\mathbf{Z}^{(1)}$.}
        \label{difference_matrix_z}
    \end{subfigure} 
    \hspace{-0.005\textwidth}
    \begin{subfigure}[b]{0.23\textwidth}
        \centering
        \includegraphics[width=\linewidth]{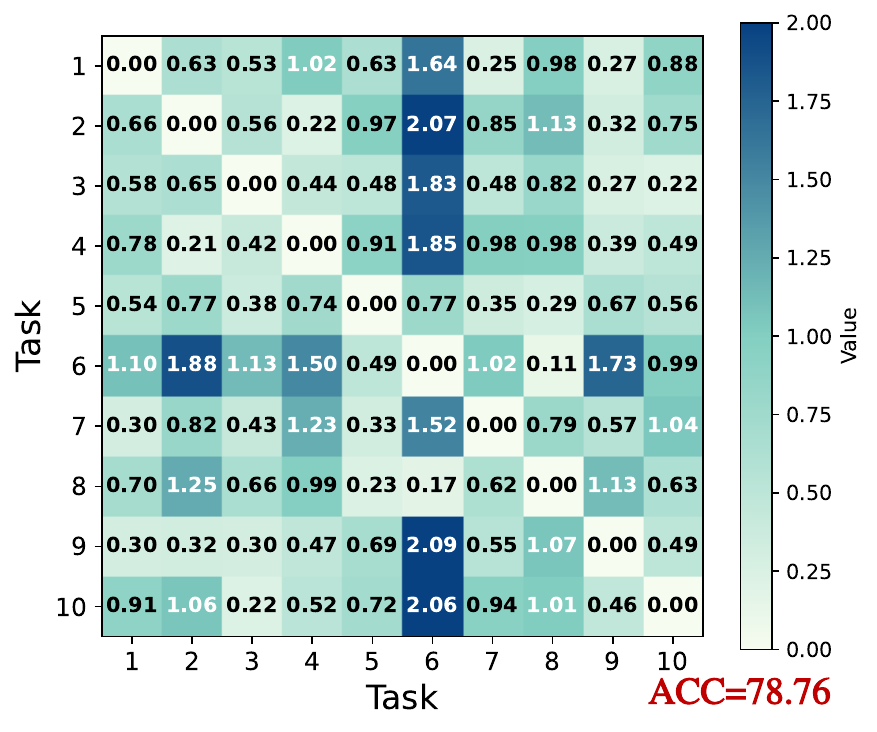}
        \caption{Difference matrix of $\mathbf{Z}_R^{(1)}$.}
         \label{difference_matrix_zR}
    \end{subfigure}
    \caption{Distributional differences in representations of the first class across ten domains on the HAR dataset.} 
    \label{difference_matrix}
\end{figure}

We further analyze the model’s performance under domain-incremental scenarios. As illustrated in Figure~\ref{difference_matrix}, after training on all domains, we use the HAR dataset as an example to examine the distributional differences of the first-class representations across domains, quantified by the KL divergence. Figure~\ref{difference_matrix}-(a) shows that the vanilla DLinear exhibits relatively low KL divergence across domains. This suggests that its representations tend to collapse toward similar distributions as training progresses, likely due to overfitting to newly encountered domains, which reduces the ability to preserve knowledge from previous domains, thereby exacerbating catastrophic forgetting. In contrast, Figure~\ref{difference_matrix}-(b) shows that our DualCD learns causal representations with more pronounced distributional differences across domains. This indicates that DualCD captures domain-specific temporal patterns while maintaining class-invariant causal structures, which facilitates better domain discrimination and more stable knowledge retention.

\subsection{Complexity Analysis} 

The computational complexity of our DualCD primarily arises from the time encoding module and the disentanglement module. The temporal encoding module can adopt any time series representation learning method. The additional computational complexity of our DualCD mainly arises from the disentanglement module, where the output and input dimensions are equal. For a clearer assessment, we evaluate two representative time series models, DLinear and PatchTST, both standalone and enhanced with our DualCD framework. First, DLinear has a computational complexity of $\mathcal{O}(L \times D) + \mathcal{O}(D^2)$, where $L$ denotes the time series length, $D$ is the feature dimension. PatchTST, which leverages patching and self-attention, involves more expensive operations with a complexity of $\mathcal{O}(P^2 \times D) + \mathcal{O}(P \times D^2)$, where $P$ is the number of patches. In contrast, DualCD incorporates only a lightweight linear module, resulting in an additional $\mathcal{O}(D^2)$ computational overhead beyond the base model. The final time complexity of DLinear+DualCD remains $\mathcal{O}(L \times D) + \mathcal{O}(D^2)$, and that of PatchTST+DualCD remains $\mathcal{O}(P^2 \times D) + \mathcal{O}(P \times D^2)$, indicating that the additional overhead introduced by DualCD is negligible.

\section{Conclusion and Discussion}
This paper addresses domain-incremental time series classification. We identify two key sources of spurious features that hinder learning of class-general representations. To counteract these, we propose DualCD, a lightweight model with a two-phase causal representation learning approach. First, a feature disentanglement module uses orthogonal masks to separate causal and spurious temporal features. Then, a dual causal intervention mechanism creates intra-class and inter-class perturbed samples to enhance robust causal representation learning. Extensive experiments validate the effectiveness of our proposed DualCD.

Although the proposed DualCD achieves accurate and efficient DI-TSC within a single dataset, it cannot be directly extended to cross-dataset scenarios due to discrepancies in the number and semantics of time series variables. In future work, we will explore variable alignment and domain-invariant representation learning to improve cross-dataset generalization.

\bibliographystyle{ACM-Reference-Format}
\bibliography{sample-base}

\end{document}